%% file: rslam_wacv19_camera_ready.tex
\documentclass[10pt,twocolumn,letterpaper]{article}

\usepackage{booktabs}
\usepackage{wacv}
\usepackage{times}
\usepackage{epsfig}
\usepackage{graphicx}
\usepackage{amsmath}
\usepackage{amssymb, comment, subfigure}
\usepackage{color}
\usepackage{enumitem,multirow}

\graphicspath{ {./figs/} }
\def\BState{\State\hskip-\ALG@thistlm}
\makeatother

\DeclareMathOperator*{\argmin}{argmin}
\include{myincludes}
\def\etal{{et al. }}
\def\lsd-slam{{LSD-SLAM }}

\usepackage[font=small,skip=2pt]{caption}
\usepackage[pagebackref=true,breaklinks=true,letterpaper=true,colorlinks,bookmarks=false]{hyperref}

\wacvfinalcopy % *** Uncomment this line for the final submission

\def\wacvPaperID{177} % *** Enter the wacv Paper ID here

\ifwacvfinal\fi
\begin{document}

%%%%%%%%% TITLE
\title{EGO-SLAM: A Robust Monocular SLAM for Egocentric Videos}

\author{Suvam Patra \quad Kartikeya Gupta \quad Faran Ahmad \quad Chetan Arora \quad Subhashis Banerjee\\
Indian Institute of Technology Delhi\\
}

\maketitle
\thispagestyle{empty}

%%%%%%%%% ABSTRACT
\begin{abstract}
Regardless of the tremendous progress, a truly general purpose pipeline for Simultaneous Localization and Mapping (SLAM) remains a challenge. We investigate the reported failure of state of the art (SOTA) SLAM techniques on egocentric videos \cite{hyperlapse, ego-seg, ego-ff}. We find that the  dominant 3D rotations, low parallax between successive frames, and primarily forward motion in egocentric videos are the most common causes of failures. The incremental nature of SOTA SLAM, in the presence of unreliable pose and 3D estimates in egocentric videos, with no opportunities for global loop closures, generates drifts and leads to the eventual failures of such techniques. Taking inspiration from batch mode Structure from Motion (SFM) techniques \cite{div_conquer,wilsonsfm}, we propose to solve SLAM as an SFM problem over the sliding temporal windows. This makes the problem well constrained. Further, as suggested in \cite{div_conquer}, we propose to initialize the camera poses using 2D rotation averaging, followed by translation averaging before structure estimation using bundle adjustment. This helps in stabilizing the camera poses when 3D estimates are not reliable. We show that the proposed SLAM technique, incorporating the two key ideas works successfully for long, shaky egocentric videos where other SOTA techniques have been reported to fail. Qualitative and quantitative comparisons on publicly available egocentric video datasets validate our results.
\end{abstract}

%Processing in sliding windows in no way makes this an offline method.

%%%%%%%%% BODY TEXT
\section{Introduction}
\begin{figure}[t]
\centering
\includegraphics[width=0.94\columnwidth]{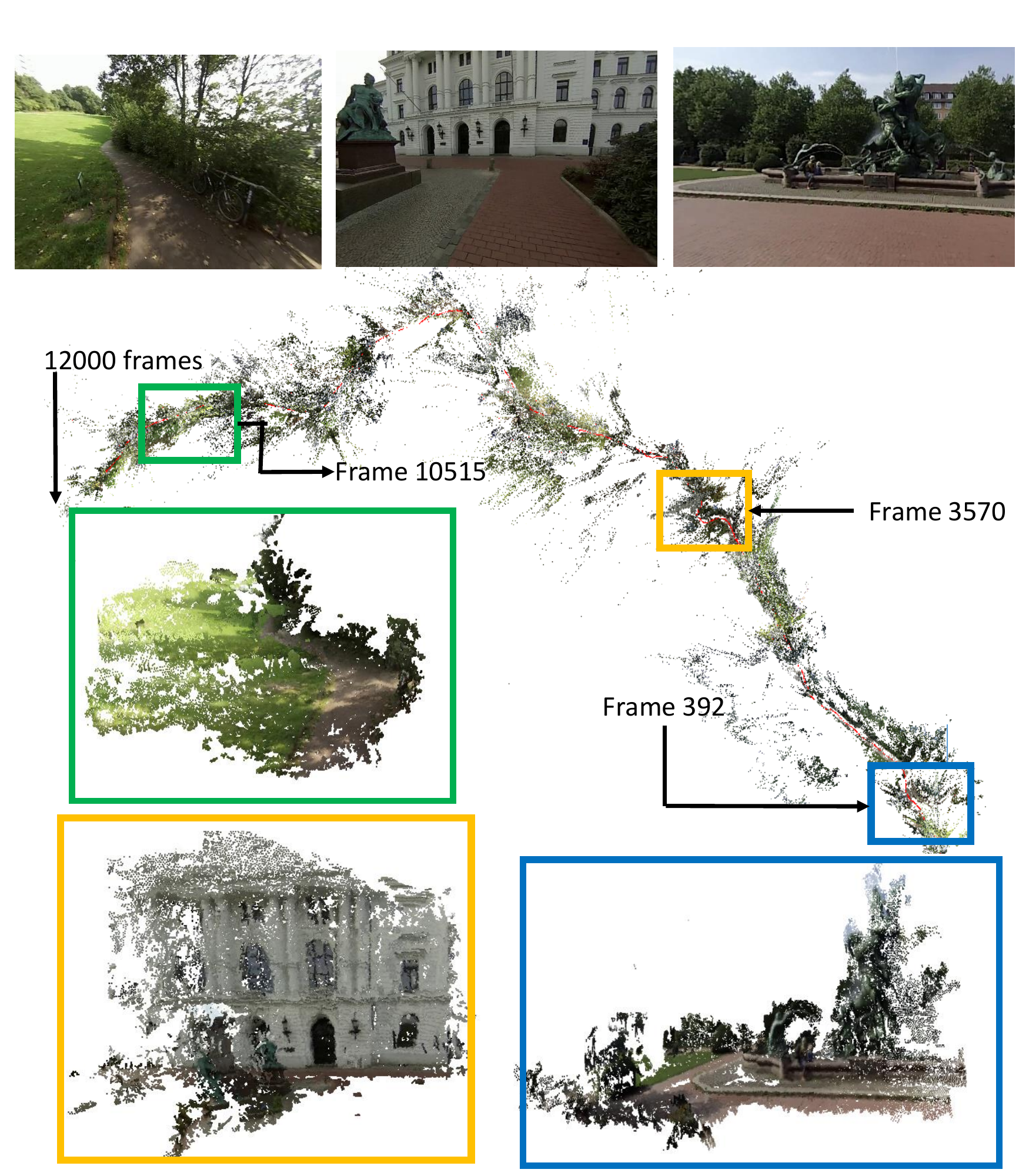}
\caption{Incremental nature of state of the art SLAM \cite{orb-slam,lsd-slam,dtslam} as well as SFM \cite{vsfm,wilsonsfm,farenzena_sfm_cviu} techniques are unsuitable for extremely unstable egocentric video when the pairwise camera pose and 3D estimates are unreliable. We propose a robust SLAM (EGO-SLAM) which solves the SLAM as an SFM problem over sliding temporal windows. The SFM problem is solved globally over the window, by first stabilizing poses using rotation and translation averaging, before going for bundle adjustment. The figure shows 3D point clouds and trajectory estimated, using the proposed algorithm, over 12000 frames from a Hyperlapse \cite{hyperlapse} $bike07$ sequence where all the other SLAM and SFM techniques have been reported to fail.}
\label{fig:motivation}
\end{figure}

Egocentric or first-person cameras \cite{google-glass, gopro, sensecam} are wearable cameras, typically harnessed on a wearer's head. First person perspective coupled with always-on nature have made these cameras popular in extreme sports, law enforcement, life-logging, home automation, and assistive vision applications \cite{ego_novelty, ego1,ego3, ego_ac_regog_4_gaze, ego_obj1, ego_ac_regco_and_chaptering, ego2, ego_ac_recog_eyecam, ego_ac_recog_2}.

Simultaneous Localization and Mapping (SLAM) has received a lot of attention from  computer vision researchers over the years. While a variety of strategies including incremental \cite{dtslam, lsd-slam,orb-slam, slam3, slam5, slam6, slam1, slam2, slam4, slam7}, hierarchical \cite{farenzena_sfm_cviu} and global  \cite{wilsonsfm} approaches have been proposed, the incremental ones remain popular because of their efficiency and scalability. Such approaches pick one frame at a time from a video stream and estimate the camera poses with respect to the 3D structure obtained so far, especially from the last few frames. Some techniques use an additional loop closure step to recognise the places visited earlier and refine the pose estimates over the loop so as to make them consistent at the intersection \cite{Williams2009}.  While the existing systems have advanced the state of the art tremendously, robustness and accuracy remain the key problems in incremental SLAM that prevent their use as general-purpose methods.

In this paper, we investigate the monocular SLAM problem with a special emphasis on {\bf EGO}centric videos, and propose a new robust {\bf SLAM} which we call {\bf EGO-SLAM}. We observe that the incremental nature of the state of the art (SOTA) SLAM techniques are unsuitable for the egocentric videos. Sharp head rotations and primarily forward motion in egocentric videos cause quick changes in the camera view and result in short and noisy feature tracks. This combined with low parallax due to dominant 3D rotation because of natural head motion of the wearer causes triangulation errors in both feature-based \cite{orb-slam} and direct techniques \cite{lsd-slam}. Relying on such 3D points causes drifts in the estimated camera trajectories leading to failure of the whole pipeline. The failure of the SOTA SLAM techniques on egocentric videos have been observed and reported by various research groups working in the area \cite{hyperlapse,ego-seg,ego-ff}. 

One of the key insights from the current work is to make the SLAM problem over egocentric videos better constrained by solving it as an SFM problem over a temporal window. Even without a careful analysis, Kopf \etal \cite{hyperlapse} also did something similar, when they carried out bundle adjustment \cite{vsfm} over large batches of 1400 frames, thereby making the problem well conditioned. However, merely doing SFM over batches is not sufficient. The idea behind batches is to make the problem better constrained, and incremental SFM techniques such as VisualSFM \cite{vsfm} over the batches defeat the core idea. In our experiments, and also reported by other research groups, \cite{hyperlapse,ego-seg,ego-ff}, batch mode visual SFM works better for egocentric videos compared to incremental SLAM, but still does not solve the problem. 
The second key insight of this paper is to suggest solving the batch SFM problem globally. We adopt the technique proposed by Bhowmik \etal \cite{div_conquer}, which first computes pairwise estimates, followed by 2D based rotation averaging, and translation averaging. The pose estimates thus obtained are fed to the bundle adjustment for joint pose and structure estimation over a temporal window. This has two advantages. First, it avoids the noisy estimates of incremental SFM. Second, motion averaging does not require  use of 3D estimates which may be erroneous at this stage. This helps to stabilize the pose estimates and obtain improved initialization for bundle adjustment (BA) which is crucial for obtaining good solutions using a BA algorithm.

%Similar to SLAM techniques, global SFM \cite{wilsonsfm} and hierarchical SFM approaches \cite{farenzena_sfm_cviu} also estimate the pose and structure of the entire image set jointly. The joint estimation, with error in poses due to shorter feature tracks and lack of parallax, is unsuitable for egocentric videos. This leads to failure of such techniques, as we show in Section \ref{sec:experiments}. Moreover, techniques such as DTSLAM \cite{dtslam} and SVO \cite{svo}, which have better stability in the presence of only rotations, also do not focus on stabilizing the pose before 3D estimation, and fail on egocentric videos. This has also been reported by various egocentric research groups \cite{hyperlapse, ego-seg, activity-rec, ego-ff}.

We note that the typical motion profiles of handheld videos, even if unstable, are not similar to that of egocentric videos. In videos obtained from handheld cameras, there are typically no dominant 3D rotations and a user often looks at the same scene from multiple viewpoints in  a scanning motion. As a result, the generated feature tracks are longer. Multiple scans give enough opportunities for loop closures making such problems simpler in comparison.

Though, the focus of this paper is on egocentric videos, the resulting framework is also suitable for other scenarios having low parallax (leading to unstable 3D estimate) and lack of loop closure opportunities, where SOTA incremental SLAM algorithms are unsuitable. One such example is videos are taken from a vehicle-mounted camera. We show that our algorithm, though not specifically designed for such situations, improves upon the SOTA on such videos as well.

The specific contributions of this paper are:
\begin{enumerate}
\item Our analysis on the failure of existing SLAM techniques for egocentric videos: We posit that computing geometry from unreliable pose estimation in an incremental fashion is the primary cause of such failures.
\item Our two key novel proposals: We suggest solving the first person SLAM as an SFM problem over temporal windows and solve each window/batch as global SFM with initialization based on 2D motion averaging. The motivation for the specific suggestions have been described above.
\item Though not specifically the focus of this paper, we also test EGO-SLAM for the vehicle-mounted cameras where similar situations exist. We show that EGO-SLAM improves the SOTA there as well.
\end{enumerate}

We contrast the proposed system to Bundler \cite{internetphoto,phototourism} and VisualSFM \cite{vsfm}, which are general purpose SFM pipelines. Our EGO-SLAM is specific SLAM pipeline for egocentric videos. Our experiments over a large set of publicly available egocentric datasets show the success, where all other SOTA SLAM algorithms have been shown to fail (Fig. \ref{fig:motivation} shows one such result). Thus, the proposed technique closes a long standing, open problem in egocentric vision and will prove to be helpful to the community.

\section{Related Work}

Based on the method of feature selection for pose estimation, SLAM algorithms from a monocular camera can be classified as feature-based, dense, semi-dense or hybrid methods. Feature-based methods, both filtering based \cite{ekfodometry} and key frame based \cite{ dtslam, ptam, orb-slam}, use sparse features like SIFT \cite{sift_2004}, ORB \cite{orb_2011}, SURF \cite{surf_2008}, etc. for tracking. The sparse feature correspondences are then used to refine the pose using structure-from-motion techniques like bundle adjustment. Due to the incremental nature of all these approaches, a large number of points are often lost during the resectioning phase \cite{triggs}.

Dense methods initialize the entire or a significant portion of an image for tracking \cite{dtam}. The camera poses are estimated in an expectation maximization framework, where in one iteration the tracking is improved through pose refinement by minimizing the photometric error, and, in alternate iterations, the 3D structure is refined using the improved tracking. To increase the accuracy of estimation, semi-dense methods perform photometric error minimization only in regions of sufficient gradient \cite{lsd-slam, semi-dense-vo-iccv-13}. However, these methods do not fare well in cases of low parallax and wild camera motions mainly because structure estimation cannot be decoupled from pose refinement.

SLAM techniques also differ on the kinds of scene being tracked: road scenes captured from vehicle mounted cameras, indoor scans from a hand-held camera, and from head-mounted egocentric cameras usually accompanied by sharp head rotations of the wearer. Visual odometry algorithms have been quite successful for hand-held or vehicle-mounted cameras \cite{lsd-slam, semi-dense-vo-iccv-13, rgbd-vo-icra, ptam, orb-slam, dtam}, but their incremental nature does not fare well for egocentric videos because of instabilities in the computation due to unrestrained camera motion, wide variety of indoor and outdoor scenes and presence of moving objects  \cite{hyperlapse,ego-seg,activity-rec,ego-ff}.

Just like SLAM, structure-from-motion (SFM) techniques can also be categorized into global, batch and incremental ones. As the names suggests, global approaches \cite{wilsonsfm} solve the global problem jointly, whereas incremental approaches like Visual SFM \cite{vsfm} inserts one frame into the estimated structure at a time. Batch mode techniques \cite{div_conquer} try to trade between the efficiency of incremental and robustness of global approaches. In recent years, the SFM techniques have seen a lot of progress, using the concepts of rotation averaging (RA) \cite{govinduefficient} and translation averaging (TA) \cite{venu01, Jiang13, MoulonMM13, wilsonsfm}. The computational cost being linear in the number of cameras, these techniques are fast, robust  and well suited for small image sets. They provide  good initial estimates for camera pose and structure using pairwise epipolar geometry, which can be refined further using standard SFM techniques.

%Loop closures in SLAM are detected using three major approaches \cite{Williams2009}: map-to-map, image-to-image and image-to-map. Clemente \etal \cite{clemente_etal_rss2007} use a map-to-map approach where they find correspondences between common features in two sub-maps. Cummins \etal \cite{CumminsIJRR08} use visual features for image-to-image loop-closures. Matching is performed based on presence or absence of these features from a visual vocabulary. Williams \etal \cite{WilliamsIROS08} use an image-to-map approach and find loop-closure using re-localization of camera by estimating the pose relative to map correspondences.

\section{Background}

The pose of  a camera $I'$ w.r.t a reference image $I$ is denoted by
a $\mathrm{3 \times 3}$ rotation matrix $\mathrm{\textbf{R} \in SO(3)}$ and a $\mathrm{3 \times 1}$ translation direction vector $\textbf{t}$.  The pairwise pose can be estimated from the decomposition of the $\textit{essential matrix}$ $E$ which binds two views using pairwise epipolar geometry such that: $E= [\textbf{t}]_\times\textbf{R}$ \cite{mvg,nisterfive}. Here $[\textbf{t}]_\times$ is a skew-symmetric matrix corresponding to the vector $\textbf{t}$. A view graph has the images as nodes and the pair-wise epipolar relationships as edges.

\subsection{Motion Averaging}\label{sec:motionavg}

Given such a view graph, embedding of the camera poses into a global frame of reference can be done using motion averaging \cite{govinduefficient,Jiang13,MoulonMM13, wilsonsfm}. The motion between a pair of cameras $i$ and $j$ can be expressed in terms of the pairwise rotation ($\textbf{R}_{ij}$) and translation direction ($\textbf{t}_{ij}$) as: $\textbf{M}_{ij} =
\left[
\begin{array}{c|c}
\textbf{R}_{ij} & s\textbf{t}_{ij}\\\hline
\mathbf{0} & 1,
\end{array}
\right]
$, where, $s$ is the scale of the translation. If $\textbf{M}_{i}$  and $\textbf{M}_{j}$ are the motion parameters of cameras $i$ and  $j$ respectively in the global frame of reference, then we have the following relationship between pairwise and global camera motions: $\textbf{M}_{ij} = \textbf{M}_{j}\textbf{M}_{i}^{-1}$.

\paragraph{Rotation averaging}: Using the above expression, the relationship between global rotations and pairwise rotations can be derived as: $\textbf{R}_{ij} = \textbf{R}_{j} \textbf{R}_{i}^{-1}$, where $\textbf{R}_{i}$ and $\textbf{R}_{j}$ are the global rotations of cameras $i$ and $j$. From a given set of pairwise rotation estimates $\textbf{R}_{ij}$,  we can estimate the absolute rotations of all the cameras  by minimising a robust sum of discrepancies between the estimated relative rotations $\bfR_{ij}$ and the relative rotations suggested by the terms $\bfR_j\bfR_i^{-1}$ \cite{govinduefficient}: $ \{\bfR_1,\cdots,\bfR_N\}=\displaystyle \argmin_{ \{\bfR_1, \cdots, \bfR_N\}} \sum_{(i,j)} \Phi \left(\bfR_j \bfR_i^{-1}, \bfR_{ij} \right)$, where $\Phi(\bfR_1,\bfR_2) = \frac{1} { \sqrt{2} }|| \log( \bfR_2 \bfR_1^{-1} )||_F$, which is the intrinsic bivariate distance measure defined in the manifold of 3D rotations on the $\mathrm{SO(3)}$. Having outlier pair­wise rotation estimates is a common problem for any rotation averaging technique. In our experiments, we have used the implementation of \cite{govinduefficient}, which handles such outliers by iterative re-weighting of the constraints using the Huber loss function.

\paragraph{Translation averaging}: The global translations $\bfT_{i}$ and $\bfT_{j}$ are related with pairwise translation directions $\bfT_{ij}$  as: $ \bft_{ij} \times (\bfT_{j} - \bfR_{ij} \bfT_{i}) = 0$. Global camera positions ($\textbf{C}_i = \bfR_i^T\bfT_i$) can be obtained as: $ \{\bfC_i\} = \argmin_{\{\bfC_i\}}{\sum_{(i,j)} d(\bfR_j^T\bft_{ij},\frac{\bfC_i -\bfC_j} {||\bfC_i-\bfC_j||})}$, where the summation is  over all camera-camera and camera-point constraints derived from feature tracks and scene 3D points. A common concern in translation averaging is to handle degeneracies arising out of linear motion. We handle the problem using additional camera-point constraints harnessed from feature tracks, and using 3D scene points as suggested in \cite{wilsonsfm}.

\subsection{Bundle Adjustment}

Triggs {\em et al.} \cite{triggs} suggests using Structure-from-Motion (SFM) to recover both camera poses and 3D structure by minimizing the following reprojection error using bundle adjustment:
\begin{equation}
\min_{\textbf{c}_j,\textbf{b}_i}\sum_{i=1}^{n}\sum_{j=1}^{m} V_{ij}D(P( \textbf{c}_j, \textbf{b}_i),\textbf{x}_{ij}\Psi(\textbf{x}_{ij}))\label{bundle1}
\end{equation}
where, $V_{ij} \in \{0,1\} $ is the visibility of the $i^{th}$ 3D point in the $j^{th}$ camera, $P$ is the function which projects a 3D point $ \textbf{b}_{i}$ onto camera $ \textbf{c}_{j}$ which is modelled using $7$ parameters ($1$ for focal length, $3$ for rotation, $3$ for position) , $\textbf{{x}}_{ij}$ is the actual projection of the $i^{th}$ point on the $j^{th}$ camera, $\Psi(\textbf{{x}}_{ij}) = 1 + r\|\textbf{{x}}_{ij}\|^2$ is the single parameter ($r$) distortion function and $D$ is the Euclidean distance. Two kinds of bundle adjustment methods are used in the literature to minimize Eq. (\ref{bundle1}). Incremental bundle adjustment technique traverses the graph  sequentially  starting from an image pair as a seed for the optimization and then keeps on adding images sequentially through resectioning of 3D-2D correspondences. The technique is used in the majority of SLAM algorithms \cite{dtslam, ptam, orb-slam}. On the other hand, Batch-mode bundle adjustment technique optimizes for all the camera poses at once by minimizing (\ref{bundle1}) globally. The approach is less susceptible to discontinuities in reconstruction or drift due to joint optimization of all cameras at once. Also, it requires an initialization for camera parameters and 3D structure which can be provided through motion averaging and linear triangulation \cite{mvg}.

\section{Proposed Algorithm}
\label{sec:algo}

\begin{figure*}[t]
\centering
\includegraphics[width=0.84\linewidth]{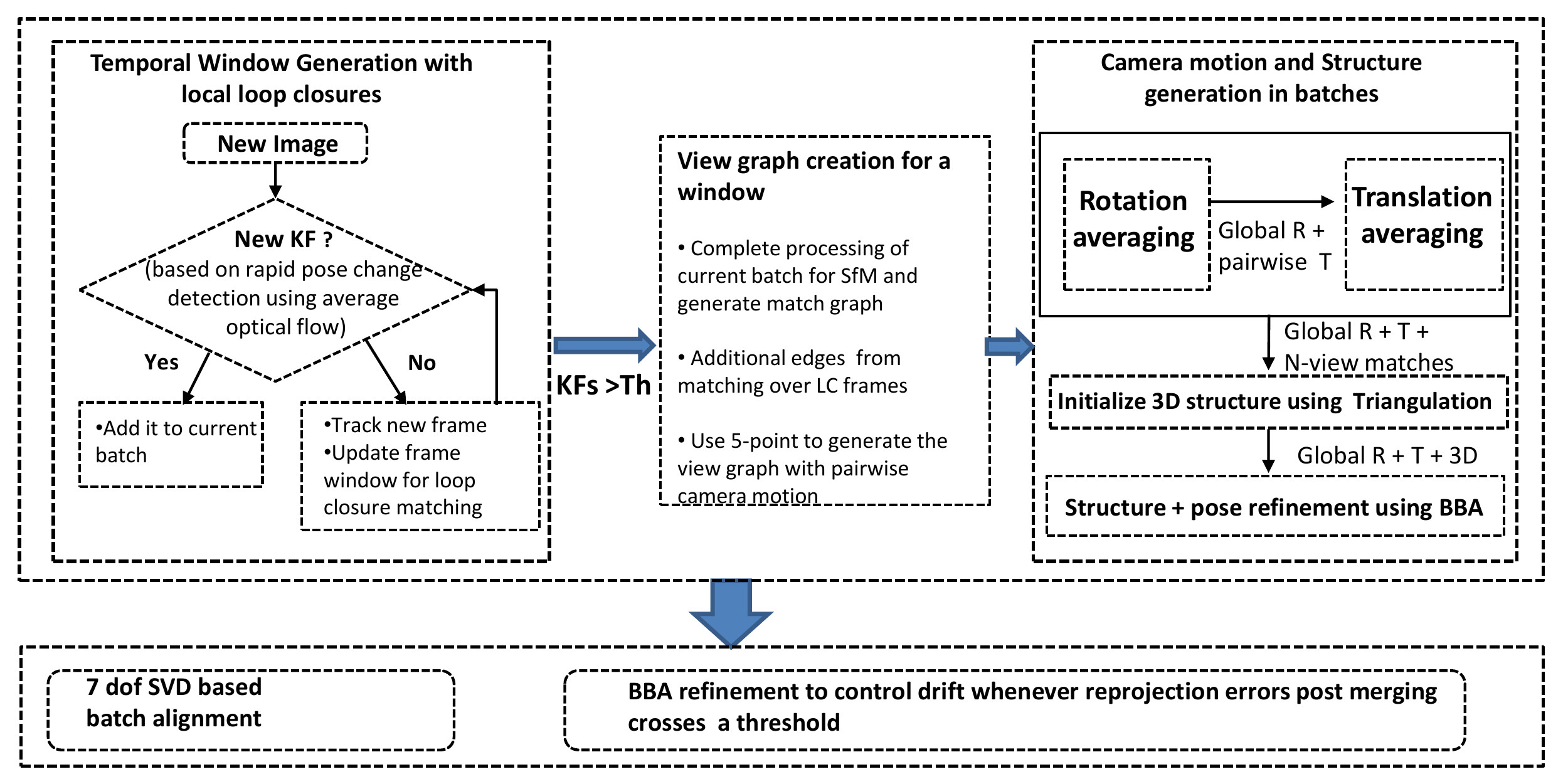}
\caption{Flow chart of the proposed EGO-SLAM}
\label{fig:framework}
\end{figure*}

\paragraph{Key Framing}

We start by processing each frame and designate a new key-frame whenever there is sufficient parallax between frames. The designation is made after every $P$ frames, or whenever the average optical flow crosses $m$ pixels, whichever happens first. This allows our method to adapt to wild motions, for example when turning. In our experiments, we typically set $P=30$ and $m=20$. We use ``good features to track'' (GFTT) for calculating optical flow. Our initial experiments with SIFT did not yield any significant benefit but increased computation time. We use bi-directional sparse iterative version of the Lucas-Kanade optical flow on intensity values of the images. Point correspondences obtained using only optical flow contain noise and therefore can create drift in the estimation of flow vectors. We filter out the correspondences which have higher bi-directional positional error.

\paragraph{Temporal Window Generation}

We do pose estimation in a temporal window of key-frames. Processing in temporal windows makes the camera motion and structure estimation problem well constrained when parallax between successive frames is small due to dominant 3D rotations, as is common in egocentric videos. We allocate a number of key-frames into a temporal window and then process each window independently. Typically each window contains around 10-30 key-frames with each key-frame separated by about 5-7 frames for the case when the wearer is walking. While lack of parallax justifies creating temporal windows, making too large a window is also problematic considering instability of motion averaging for large windows. A smaller window size also helps in controlling drifts and breaks in structure estimation.

\begin{figure}[t]
\centering
\subfigure[]{\label{fig:batches-incremental}
\includegraphics[width=0.46\linewidth]{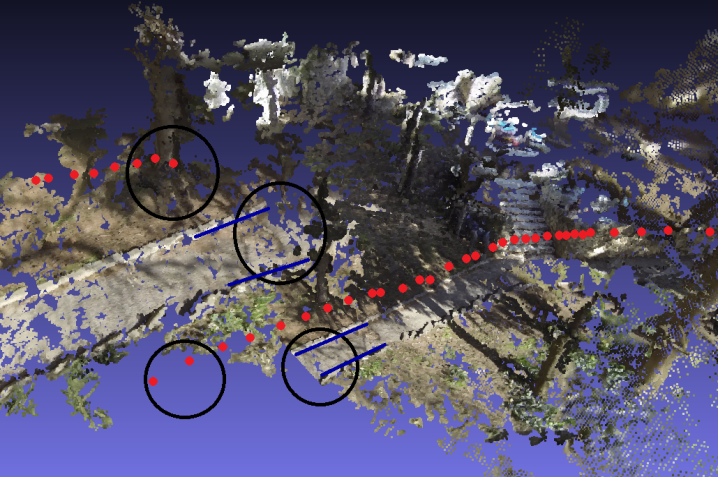}} \hspace{0.01cm}
\subfigure[]{\label{fig:batches-us}
\includegraphics[width=0.46\linewidth]{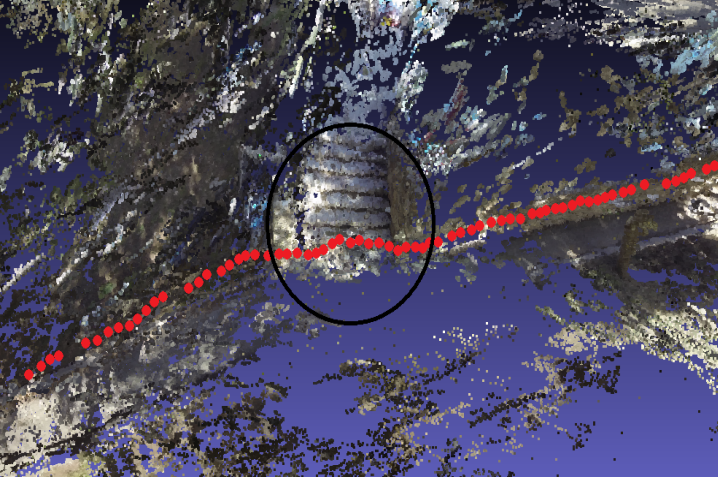}} \\
\subfigure[]{\label{fig:batches-large-batch}
\includegraphics[width=0.46\linewidth]{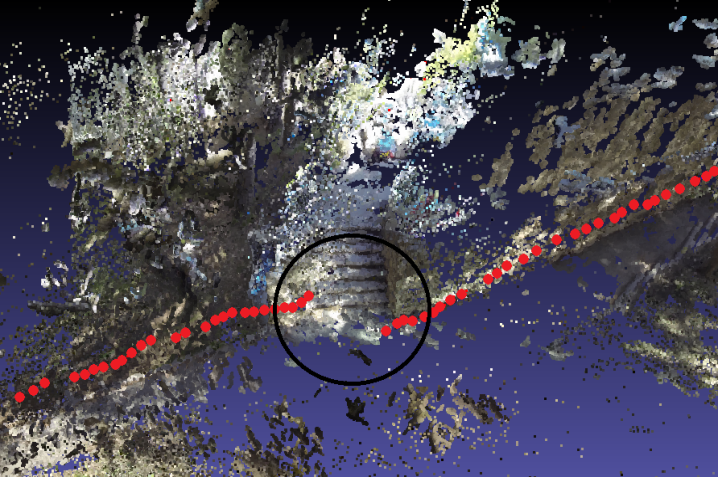}}
\hspace{0.01cm}
\subfigure[]{\label{fig:batches-large-batch-refimg}
\includegraphics[width=0.46\linewidth]{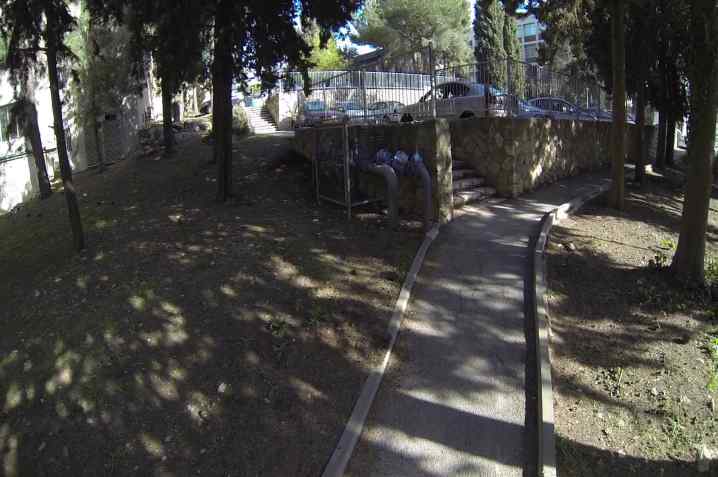}}
\caption{Incremental SLAM is problematic for egocentric videos due to lack of parallax between successive frames. We propose batch mode processing to stabilize the trajectory estimation first. (a), (b) and (c) show output with a batch size of 1, 30 and 500 respectively. (d) is the reference image. Large batch size may also cause problems in motion averaging convergence and breaks in SFM causing trajectory break highlighted in (c), structure error as well as trajectory break, highlighted in (a) but corrected by small batch size in (b). The sequence is taken from HUJI dataset \cite{ego-seg}.}
\label{fig:batches}
\end{figure}

It may be noted that if we choose a window size of $1$, our method essentially becomes incremental strategy as is common in state of the art methods like VisualSFM \cite{vsfm}. In Figure \ref{fig:batches}, we analyze the effect of window size on trajectory estimation for a window size of 1, 30 and 500. The estimation process fails for 1 and 500 but works best for 30 in this particular case.

In global SFM, densely connected entire view graph allows larger window sizes to yield more tightly bound constraints for motion averaging and hence better estimates. However, in our case feature correspondences are generated transitively through sequential tracking, which leads to drift in larger windows due to poor estimation of pairwise epipolar geometry for redundant paths (longer edges).

Though, most of the results in this paper have been produced with non-overlapping batches, in case a lower latency is required, one can use sliding windows with significant overlaps as well.

\paragraph{Local Loop Closures}

We use local loop closures to handle large rotations in our input videos. This gives extra constraints for stabilising the camera estimates. Global loop closure is an important step in traditional SLAM to fix the accumulated errors over pose estimation. However, in case of the egocentric videos, where the motion of the wearer is linear forward, a user may not revisit a particular scene point for long, making global loop closure sometimes impossible. Also, given the wild nature of egocentric videos, the camera poses and trajectories tend to drift quickly unless fixed by loop closures immediately. We observe that in a natural walking style, a wearer's head typically scans the scene left to right and back. The camera looks at the same scene multiple times, thus providing opportunities for a series of short local loop closures. We take advantage of this phenomenon by using local loop closures Patra \etal \cite{wacv17_egomotion}, to improve the accuracy of the estimated camera poses.

\begin{figure}[t]
\centering
\subfigure{\includegraphics[width=0.29\linewidth]{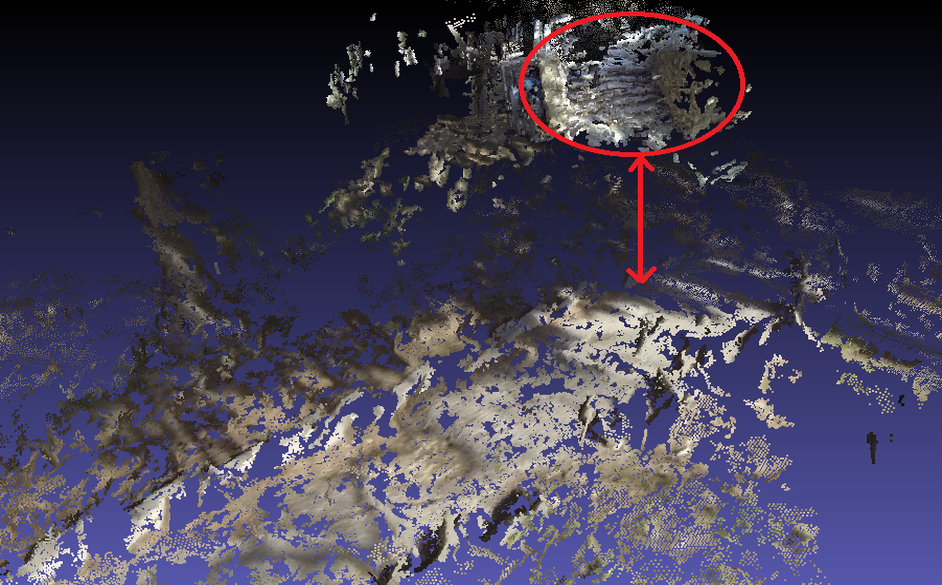}}\hspace{0.4cm}
\subfigure{\includegraphics[width=0.29\linewidth]{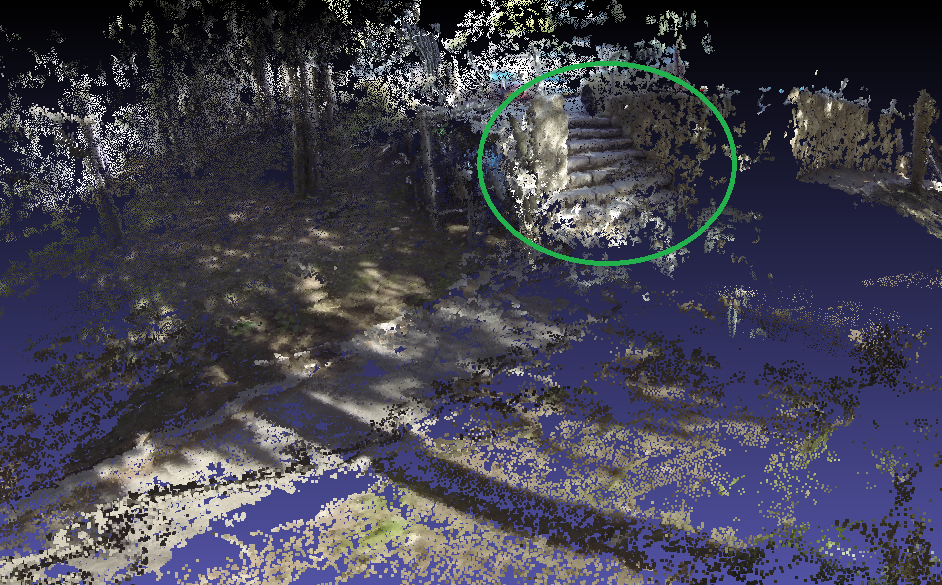}}\hspace{0.4cm}
\subfigure{\includegraphics[width=0.29\linewidth]{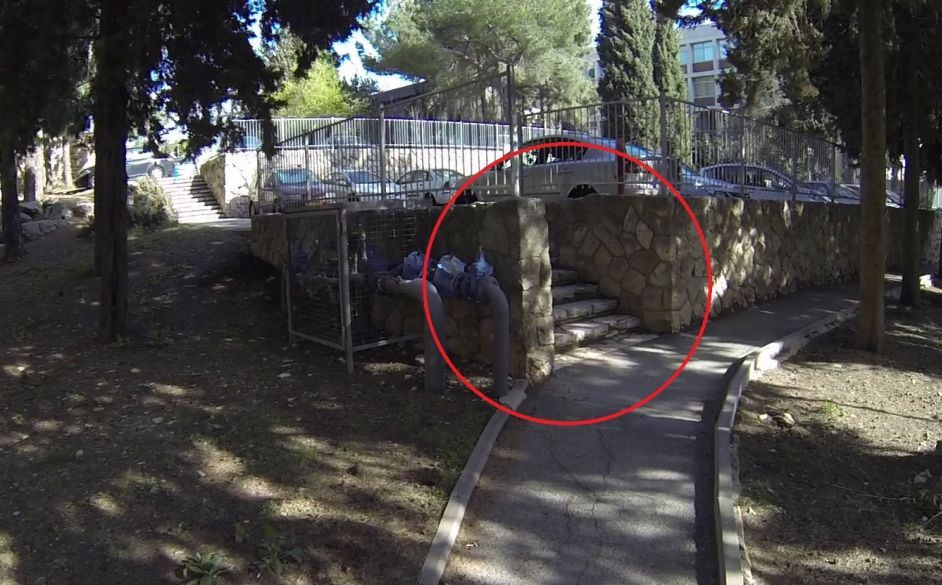}}\\
\caption{Loop closures are an important step in a SLAM algorithm but may never be applied in an egocentric video because of usual forward motion of the wearer. In this paper, we have suggested local loop closures for egocentric videos. First and second images show structure estimation without and with local loop closures respectively. Third image is the reference view. Note the `hanging' stairs in the first image without loop closure.}
\label{fig:loop-closure}
\end{figure}

We maintain a set of last few key-frames and when considering a new key-frame, we estimate its pairwise pose with these existing key-frames to discover redundant paths. The additional edges added in the view graph during this stage helps in local loop closures through motion averaging as described in the following step. Figure \ref{fig:loop-closure} shows the effect of loop closures on the estimation. In the absence of these extra edges and thus the local loop closures, the structure around the staircase gets deformed in the scale and also shifts above the ground, thus causing breaks.

\paragraph{Camera Pose Estimation}

After adding extra edges in the view graph for local loop closure, we use the five-point algorithm \cite{nisterfive} for estimating the pair-wise epipolar geometry for a batch. This provides sufficient constraints for motion averaging. We first use rotation averaging for finding global rotation estimates followed by translation averaging. Note that providing a good initial estimate of camera pose, without using any 3D structural information, is essential for any SLAM approach to work successfully on egocentric videos. This is where the proposed approach is critically different from state of the art approaches.

After robustly estimating rotations, we use a mixture of two different methods for averaging the translations. To initialize the global translations we generate an initial guess using global convex optimization technique specified in \cite{ozyesil2014robust_cvpr} and subsequently refine the solution using the approach of \cite{wilsonsfm}. This provides an excellent initial estimate for the camera poses. During this phase, the camera intrinsics remain constant.

\paragraph{3D Structure Estimation}

Once the camera poses are robustly initialized, the 3D structure is setup using linear triangulation as specified in \cite{mvg}. We further refine the initial structure and camera poses using a final run of Window mode Bundle Adjustment (WBA). It estimates all the camera poses and 3D points simultaneously using Bundle adjustment \cite{triggs}. The convergence of bundle adjustment is very fast due to the good initialization as described above. Also, this phase allows us to refine camera intrinsics through WBA.

\paragraph{Merging and WBA Refinement with Resectioning}

As the last step we merge the structure obtained from successive temporal windows using 7 {\em dof} alignment based on SVD as suggested in \cite{umeyama}. During the merging step, new points which were not used previously due to not being visible in most cameras, are added back, as these points get stable at this stage with more cameras viewing them now. A final round of global BBA based refinement is run in the background whenever the cross batch reprojection errors get high. This leads to a non-linear refinement in the scale of the estimated structure and poses. We describe the complete algorithm of EGO-SLAM in Figure \ref{fig:framework}.

\begin{figure}[t]
\centering
\subfigure{\includegraphics[width=0.32\linewidth]{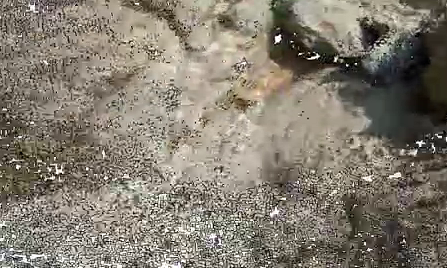}}
\subfigure{\includegraphics[width=0.32\linewidth]{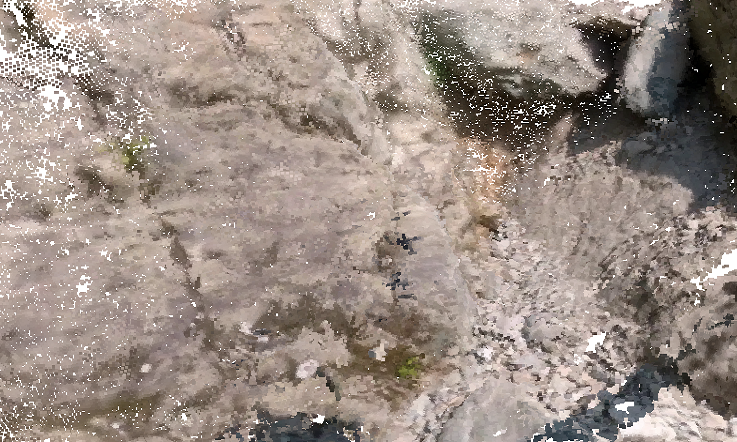}}
\subfigure{\includegraphics[width=0.32\linewidth]{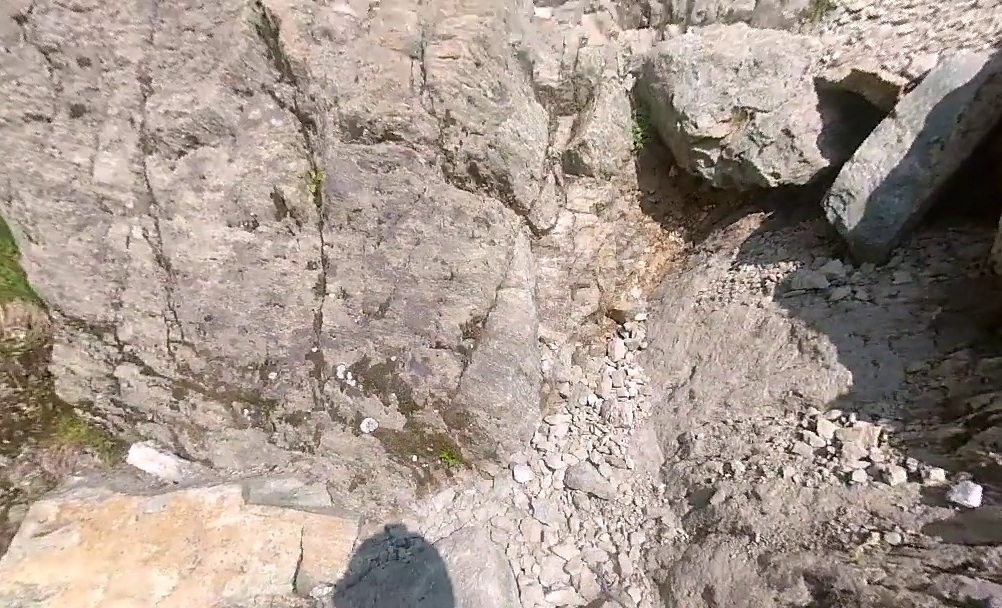}}
\caption{Comparison of the estimated structure on a challenging Hyperlapse $climbing03$ sequence \cite{hyperlapse}. State of the art SLAM fails here and authors of hyperlapse have reported using SFM algorithm by manually dividing the sequence into batches of 1400 frames. EGO-SLAM works without fail on the complete sequence. Left: Dense depth map generated by \cite{hyperlapse} using CMVS \cite{cmvs}. Middle: Corresponding dense depth map generated by EGO-SLAM. Right: A reference view}
\label{fig:hyperlapse}
\end{figure}

\section{Experiments and Results}
\label{sec:experiments}

In this section, we validate the robustness of EGO-SLAM on various publicly available video datasets captured from egocentric, handheld and vehicle mounted cameras. We have implemented portions of our algorithm in C++ and MATLAB. All the experiments have been carried out on a regular desktop with Core i7 2.3 GHz processor (containing 4 cores) and 32 GB RAM, running Ubuntu 14.04.

Our algorithm requires the intrinsic parameters of the cameras for SFM. For sequences taken from public sources, we use the calibration information based on the make and the version of the cameras provided on their websites.

Note that various egocentric research groups have reported the failure of various SLAM methods on egocentric videos. Therefore, we have restricted our attention to comparison with latest SLAM techniques which have been published after those reports: mainly ORB-SLAM, but also LSD-SLAM and PTAM for indicative purposes.

For visual clarity, we show the dense 3D map in all our examples by carrying out the dense reconstruction of some portions using CMVS \cite{cmvs}. We provide to CMVS  the camera poses and the sparse structure computed using our algorithm. Note that CMVS can produce high quality output only if the pose and the initial structure estimates are correct, and this also serves as a test for our results. We present results with views of the point clouds from a single vantage point w.r.t. the reference image. For more views from better vantage points please refer to our supplementary video. 

Please note that in our implementation, we use optical flow for image matching because of simplicity and speed. However, our pipeline does not preclude the use of feature descriptor based matching for relocalization and mapping applications (see Section \ref{sec:reloc} for a discussion). For a video of $1280\times 800$ at 60fps with a batch size of 20 Key Frames (KF), on an average a batch lasts for 0.8 - 2 sec based on the type of video (usually shorter for an egocentric and longer for a car mounted video). Some indicative timings for a set of 49 frames from which 20 frames were chosen as KFs are: Relative Pose: 10.71 sec; Motion Averaging: 1.34 sec; Triangulation: 1.66 sec; BBA: 0.13 sec; and Batch Merging: 3 sec. Note that our code is unoptimized. E.g., Finding relative pose have been shown to work in real time by others.

\begin{figure}[t]
\centering
\subfigure{\includegraphics[width=0.32\linewidth]{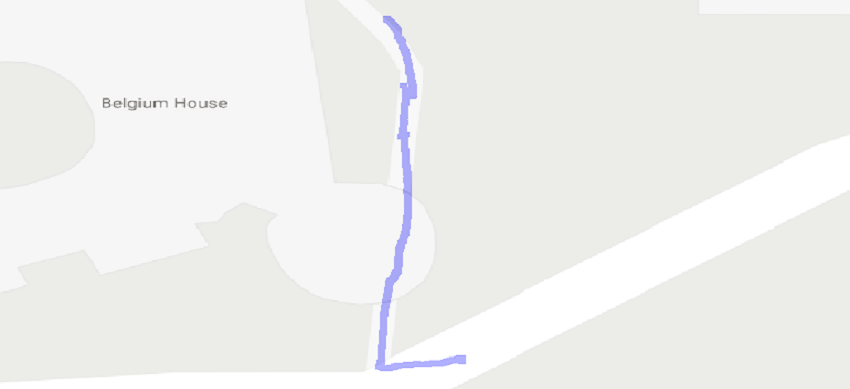}}
\subfigure{\includegraphics[width=0.32\linewidth]{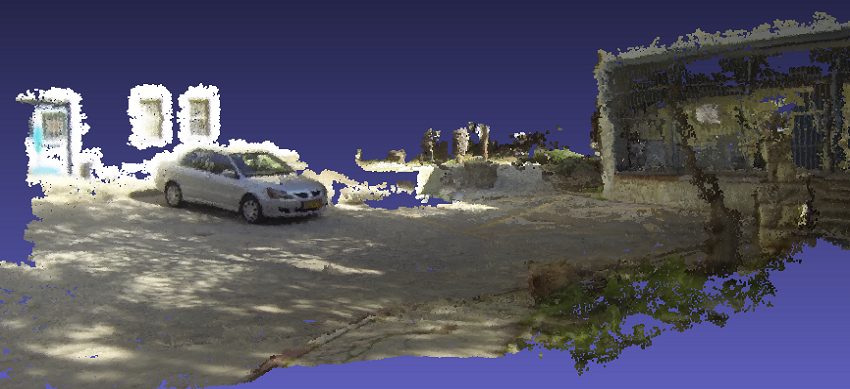}}
\subfigure{\includegraphics[width=0.32\linewidth]{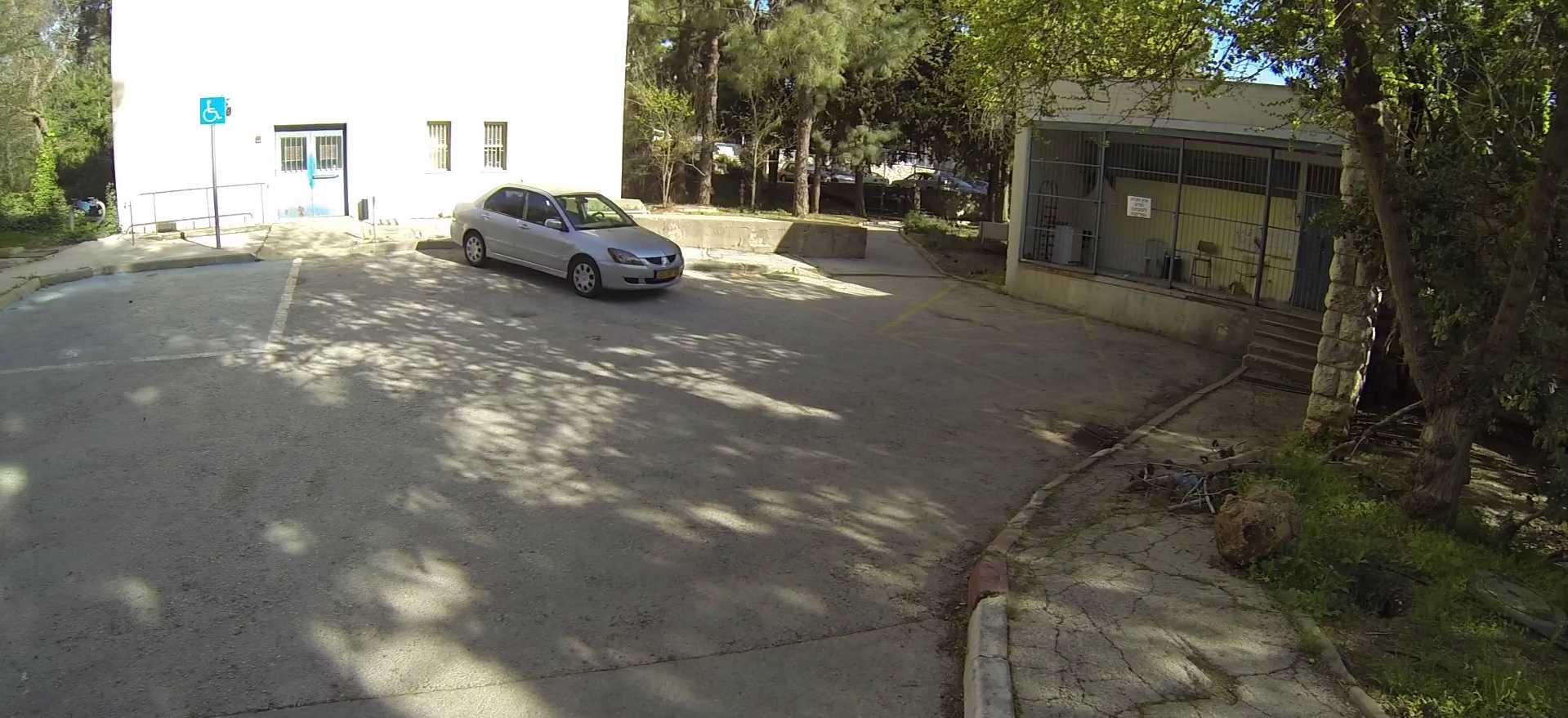}}
\caption{Our result on another challenging sequence from the HUJI data set \cite{ego-seg}. Here the wearer is walking in a narrow alley and even makes a sharp 360 degree turn. Left: Estimated trajectory on superimposed on Google map. Middle: Dense depth map of a portion obtained using CMVS \cite{cmvs}. Right: Reference View}
\label{fig:huji}
\end{figure}

\begin{table}[t]
\setlength{\tabcolsep}{4pt}
\begin{center}
\begin{tabular}{lcccc}
\toprule[1.5pt]
\multirow{2}{*}{{\bf Seq. Name}} & \multirow{2}{*}{{\bf \#Frames}} & \multicolumn{3}{c}{\bf \#Breaks in the Seq.} \\ \cmidrule{3-5}
& & {EGO} &{ORB}\cite{orb-slam}&{LSD}\cite{lsd-slam}\\ \midrule
$Bike07$ \cite{hyperlapse}& 12000 & 0 & 14 & 17 \\
$Climbing03$ \cite{hyperlapse}& 3866 & 0 & 4 & 20 \\
$Yair\_5$ \cite{ego-seg}& 3634 & 0 & 2 & 2 \\
$Yair\_1\_p2$ \cite{ego-seg}& 3601 &0 & 1 & 2 \\
$Yair\_6$ \cite{ego-seg}& 3357 &0 & 1 & 1 \\
\bottomrule[1.5pt]
\end{tabular}
\end{center}
\caption{Number of breaks suffered by various methods on 5 videos from the Hyperlapse \cite{hyperlapse} and the HUJI egoseg \cite{ego-seg} datasets}
\label{tablebreaks}
\end{table}

\subsection{Egocentric Videos}

We have tested EGO-SLAM on various Hyperlapse sequences \cite{hyperlapse}. The {\em bike07} \cite{hyperlapse} video in the dataset is a very challenging sequence with wild head motions, fast forward movements, and sharp turns. Both \cite{dtslam,orb-slam} break on this sequence. We have already shown the computed trajectory for the sequence in Figure \ref{fig:motivation}. In the same figure, we have shown the 3D map by carrying out dense reconstruction of some portions using CMVS \cite{cmvs} based on the camera poses and the sparse structure computed using our algorithm. In Figure \ref{fig:hyperlapse} we compare the dense 3D structure of a portion computed using EGO-SLAM with the one given in Hyperlapse. It is to be noted that in \cite{hyperlapse}, pose and 3D structure are computed using SFM over batches of 1400 frames.

We present similar results on a similarly challenging $Huji\_Yair\_5$ sequence from the HUJI EgoSeg dataset in Figure \ref{fig:huji}. All the state-of-the-art SLAM techniques have been reported to fail on these datasets \cite{ego-seg,activity-rec,ego-ff}.

One of our claims in this paper is on the robustness of proposed method over the SOTA. To validate this claim, we count the number of breaks/crashes suffered by various methods while processing egocentric videos from Hyperlapse \cite{hyperlapse} and HUJI EgoSeg \cite{ego-seg} datasets. Table \ref{tablebreaks} shows the number of such breaks.

\begin{figure}[t]
\centering
\subfigure[]{\includegraphics[width=0.44\linewidth]{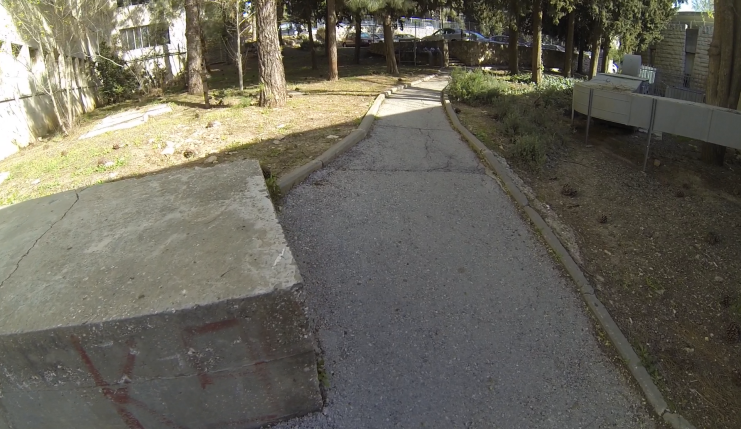}}\hspace{0.2cm}
\subfigure[]{\includegraphics[width=0.44\linewidth]{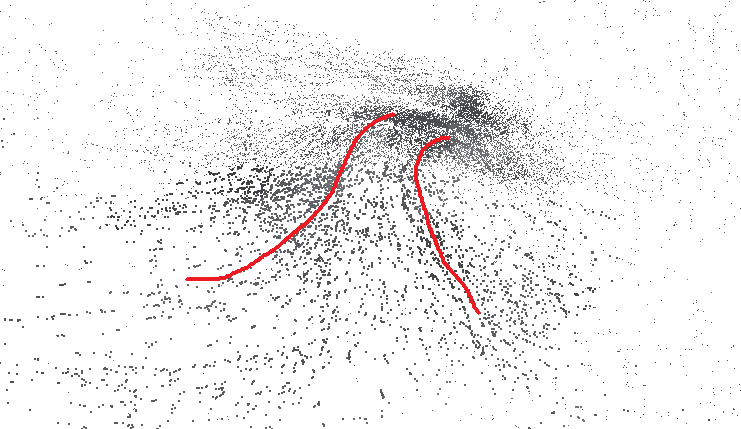}} \\
\subfigure[]{\includegraphics[width=0.44\linewidth]{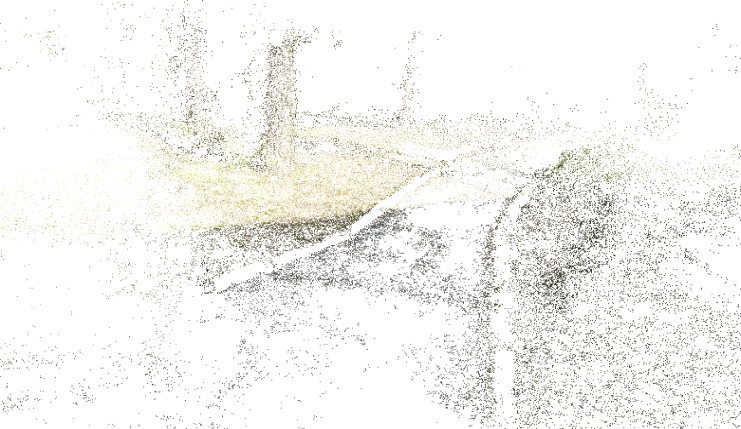}}\hspace{0.2cm}
\subfigure[]{\includegraphics[width=0.44\linewidth]{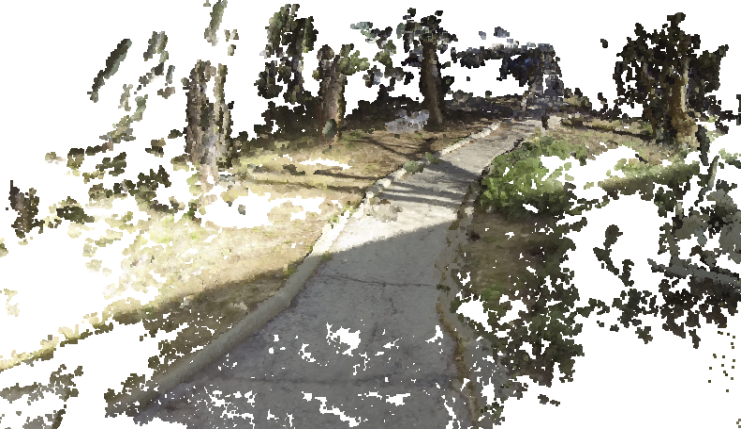}}\\
\caption{Poor 3D estimation by SOTA is one of the major reasons of breaks. (a) shows a reference view from a HUJI sequence \cite{ego-seg}, (b) poor structure estimation by ORB-SLAM (road highlighted) just before the break (c) shows the correct structure estimation by EGO-SLAM at the same point  which is made dense by CMVS \cite{cmvs} in (d).}
\label{fig:orb_vers_ours}
\end{figure}

\begin{table}[t]
\begin{center}
\begin{tabular}{lcccc}
\toprule[1.5pt]
{\bf Motion} & {\bf Traj. len.} &\multicolumn{3}{c}{\bf RMSE (cm) of 3D} \\ \cmidrule{3-5}
{\bf profile}& {\bf (m)} & {EGO} &{ORB}\cite{orb-slam} & {LSD}\cite{lsd-slam}\\ \midrule
$frontal$ & 2.0 & \textbf{20.6} & 30.5 & 39.8 \\
$left-right$ & 4.0 & {\bf 24.9} & 26.6 & 35.0 \\
$egomotion$ & 3.7 & {\bf 22.7} & 25.4 & 47.9 \\
\bottomrule[1.5pt]
\end{tabular}
\end{center}
\caption{Accuracy analysis of estimated structure and comparison with SOTA using a synthetic scene and different motion profiles.}
\label{tablesynth}
\end{table}

There are no benchmark datasets with ground truth trajectory or structure for egocentric videos. Therefore, we created a synthetic setup for quantitative error estimation. We created a synthetic scene with different planes of various sizes (max $5.12m\times 5.12m$) at different depths. Since the depths are known we now use the projected images under different motion profiles: frontal, left-right, and egocentric to estimate the 3D using LSD-SLAM, ORB-SLAM  and EGO-SLAM, and then compare the estimated depths from each of these methods against the ground truth. We present this analysis in Table \ref{tablesynth}. More details and visualization can be found in supplementary.

%Since the trajectory length (in distance covered) information is not available for these videos, we only show the number of frames in the sequence. We have captured additional videos of our own with length information as well. Due to the paper length constraints, we show the break statistics with trajectory length on these videos in the supplementary material.

Note that the breaks suffered by the SOTA are due to 3D tracking failures due to poor localization and inferred structure. We confirm this claim and show in Figure \ref{fig:orb_vers_ours}  the incorrect structure computed before the break by  ORB-SLAM. We also show the corresponding correct structure computed by our method in the same figure for comparison.

We also compare our method with Hierarchical SFM (HSFM) \cite{farenzena_sfm_cviu}. We ran the free version of their commercial software (Zephyr Lite \cite{zephyr}) on bike07 sequence, and found it to break first time at around frame no 1927 and after that it suffered from breaks at multiple places due to resectioning (5 times within the first 7000 frames) whenever there were sharp turns. The structure estimated was also deformed.

\begin{table}[t]
\begin{center}
\begin{tabular}{llcc}
\toprule[1.5pt]
\multirow{2}{*}{{\bf Seq.}} & {\bf Dimension} & \multicolumn{2}{c}{\bf Trajectory RMSE (m)} \\
\cmidrule{2-4}
 & $m \times m$ & EGO & ORB\cite{orb-slam} \\
 \midrule
 KITTI 00 & $564\times496$ & 3.34 & 6.68  \\
 KITTI 01 & $1157\times1827$ & 67.09 & X  \\
 KITTI 02 & $599\times946$ & 7.75 & 21.75  \\
 KITTI 03 & $471\times199$ & 0.44 & 1.59  \\
 KITTI 04 & $0.5\times394$ & 1.75 & 1.79  \\
 KITTI 05 & $479\times426$ & 3.85 & 8.23  \\
 KITTI 06 & $23\times457$ & 11.63 & 14.68  \\
 KITTI 07 & $191\times209$ & 2.41 & 3.36  \\
 KITTI 08 & $808\times391$ & 5.87 & 46.58  \\
 KITTI 09 & $465\times568$ & 6.97 & 7.62  \\
 KITTI 10 & $671\times177$ & 0.85 & 8.68  \\
 \bottomrule[1.5pt]
 \end{tabular}
\end{center}
\caption{Our results on videos taken from vehicle mounted cameras on the KITTI dataset \cite{KITTI}. RMS error of computed trajectories (in meters) with respect to the ground truth trajectory show that we improve upon the SOTA on such videos as well. ``X'' denotes failure in estimation. Visualization of some of the computed trajectories is in the supplementary.}
\label{tablermseKITTI}
\end{table}

\subsection{Vehicle Mounted Cameras}

Though, the focus of this paper is on egocentric videos, our algorithm is equally applicable for other capture scenarios where there is low parallax between consecutive frames and lack of global loop closure opportunities. One such case arises from vehicle mounted forward looking cameras. We have experimented with one such challenging dataset \cite{KITTI}.
%Figure \ref{fig:kitti} shows the trajectories computed using our algorithm along with ground truth trajectories on various sequences from the dataset.

Table \ref{tablermseKITTI} shows the RMS error of the computed trajectory with respect to the ground truth. Comparison with a state of the art method, ORB-SLAM \cite{orb-slam}, indicates that we perform better on such videos as well. Note that LSD-SLAM \cite{lsd-slam} does not work on the KITTI videos.

\subsection{Handheld Cameras}

\begin{figure}[t]
\centering
\includegraphics[width=0.42\linewidth]{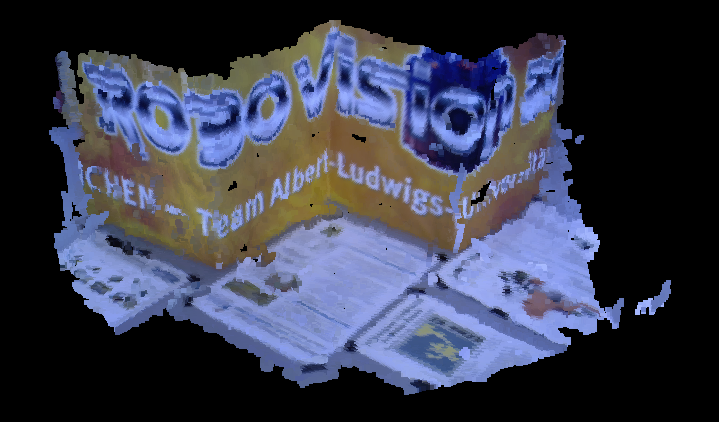}\vspace{0.4cm}
\includegraphics[width=0.42\linewidth]{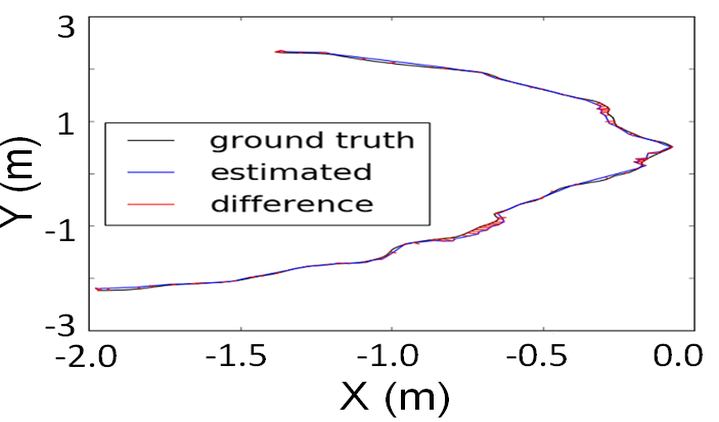}
\caption{Left: Dense depth map computed using EGO-SLAM + CMVS \cite{cmvs} on $fr3\_str\_tex\_far$ seq. (TUM dataset \cite{tum}). Right: Comparison with ground truth  trajectory after 7 dof alignment}
\label{fig:scanning2}
\end{figure}

\begin{table}[t]
\renewcommand*{\arraystretch}{0.8}
\begin{center}
\begin{tabular}{lcccc}
\toprule[1.5pt]
\multirow{2}{*}{{\bf Seq.}} & \multicolumn{4}{c}{\bf RMSE (cm) of Trajectory} \\ \cmidrule{2-5}
& EGO & {ORB}\cite{orb-slam} & {PTAM}\cite{ptam} & {LSD}\cite{lsd-slam}\\ \midrule
f1\_fl & \textbf{1.60} & 2.99 & X & 38.07 \\
f1\_d & {\bf 1.34} & 1.69 & X & 10.65 \\
f3\_l\_off & {\bf 1.03} & 3.45 & X & 38.53 \\
f3\_s\_t\_f & 1.06 & {\bf 0.77} & 0.93 & 7.95 \\
f3\_ns\_t\_f & 13.89 & X & 4.9 / 34.7 & 18.31 \\
f3\_s\_t\_n & {\bf 1.03} & 1.58 & 1.04 & X \\
f3\_ns\_t\_n & {\bf 1.31} & 1.39 & 2.74 & 7.54 \\
\bottomrule[1.5pt]
\end{tabular}
\end{center}
\caption{Comparison of RMS error with respect to ground truth trajectory on a few sequences from the TUM dataset \cite{tum} of handheld video. Our error is better than LSD-SLAM on these sequences and also better than ORB-SLAM and PTAM in most cases. ``X'' denotes failure in estimation. Detailed information on the sequences can be found in the supplementary material.}
\label{tablermseTUM}
\end{table}

We emphasize that our technique is specially geared for egocentric videos. For hand held videos our technique holds no special advantage and only works as well as traditional SLAM algorithms. However, to confirm the applicability of the proposed technique as a generic SLAM, we provide experimental results on a few hand held benchmarks as well.

We have used the TUM Visual odometry dataset \cite{tum} for the analysis of videos captured from handheld cameras. Figure \ref{fig:scanning2} shows the dense reconstruction and the trajectory estimated by the proposed method. Note that the graph shown in the figure also contains the ground truth trajectory, but the estimated trajectory is completely aligned with the ground truth and hides it completely.

The TUM dataset also allows us to compute the RMS error of the computed trajectory with respect to the ground truth trajectory. Table \ref{tablermseTUM} shows the error for EGO-SLAM as well as the ones reported by the other SOTA techniques on the same sequence. We match and often improve the state of the art even for regular hand-held videos as well. Note that, for the fr3\_nostructure\_texture\_far sequence ORB-SLAM fails due to planar ambiguity and PTAM produces ambiguous results due to different initializations every time. Hence for this case PTAM also produces unreliable results.

\subsection{Relocalization}
\label{sec:reloc}

\begin{figure}[t]
\centering
\includegraphics[width=0.7\columnwidth]{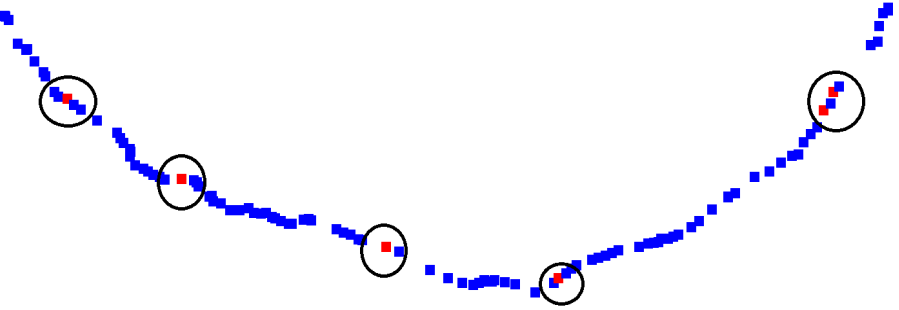}
\caption{Our pipeline can also use standard feature descriptors for relocalization. The figure shows localized novel cameras on the precomputed trajectory using our method (see paper text for details). The estimated locations (red dots) near the trajectory indicate successful localization in TUM fr3\_str\_tex\_far sequence.}
\label{fig:reloc}
\end{figure}

\begin{table}[t]
\renewcommand*{\arraystretch}{0.8}
\begin{center}
\begin{tabular}{lcccc}
\toprule[1.5pt]
\bf Method   & \multicolumn{2}{p{2.5cm}}{\bf Rotation (deg.)}& \multicolumn{2}{p{2.5cm}}{\bf Position (cm)} \\
\cmidrule{2-5}
& $Mean$&$Median$ &$Mean$ &$Median$ \\
 \midrule
 Without BA &0.0198 &0.0216 & 1.004 & 1.040 \\
 With BA &0.0062 &0.0051 & 0.975 &0.977 \\
\bottomrule[1.5pt]
\end{tabular}
\end{center}
\caption{Quantitative analysis of relocalization error. We perform relocalization as shown in Figure \ref{fig:reloc} and compute error in camera rotation (degrees) and absolute position (cm) after relocalization for novel frames. Smaller error indicates successful localization.}
\label{tablereloc}
\end{table}

Relocalization error is a popular metric to measure the accuracy of estimated 3D structure. In EGO-SLAM, we use optical flow for image matching for the sake of simplicity and speed. Since optical flow vectors do not have associated feature descriptors, they cannot be used for relocalization and mapping. However, our pipeline does not preclude use of such feature descriptors for relocalization.

To demonstrate relocalization using our framework, we train a vocabulary tree \cite{vocabnister} using the SIFT \cite{sift_2004} features computed from the key-frames in the TUM $fr3\_str\_tex\_far$ sequence \cite{tum}. We then use a set of frames which are not key-frames to calculate the relocalization error. We carry out feature matching with the key-frames using vocabulary tree, reject outliers using the pre-computed trajectory of the key-frames, and estimate the pose of the unknown frames using 3D-2D correspondences.

In Figure \ref{fig:reloc}, we plot the relocalized unknown frames on the computed trajectory. Location of the frames on the trajectory indicate the correctness of relocalization. In Table \ref{tablereloc} we show the accuracy of relocalization with respect to the ground-truth  both with and without a final BA refinement.

It may be noted that relocalization also facilitates global loop closures in  our original match graph.

\section{Conclusion}

Despite tremendous progress made in recent SLAM techniques, running such algorithms for many categories of videos still remain a challenge. We believe that careful case by case analysis of such challenging videos may provide crucial insights into improving the SOTA. Egocentric a.k.a first person videos are one such category we focus on in this paper. We observe that incremental estimation employed in most current SLAM techniques often cause unreliable 3D estimates to be used for trajectory estimation. We suggest to first stabilize the trajectory using 2D techniques and then go for structure estimation. We also exploit domain specific heuristics such as local loop closures. Interestingly, we observe that the proposed technique improves the SOTA for videos captured from vehicle mounted cameras also. Finally, many applications like hyperlapse \cite{hyperlapse}, and first person action recognition \cite{ego_ac_recog_2,first-person-action} could have been solved by principled camera pose and structure estimation. But the authors of such works were forced to take other approaches because of inability of SOTA SLAM techniques on egocentric videos. We believe many such current and future researchers will benefit by the use of proposed technique.\\
{\bf Acknowledgement:} This work was supported by a research grant from Continental Automotive Components (India) Pvt. Ltd.

{\small
\bibliographystyle{ieee}
\bibliography{egbib}
}

\end{document}

%% file: myincludes.tex
\newcommand{\ba}{\begin{array}}
\newcommand{\ea}{\end{array}}

\newcommand{\bc}{\begin{center}}
\newcommand{\ec}{\end{center}}
\newcommand{\bit}{\begin{itemize}}
\newcommand{\eit}{\end{itemize}}

\newcommand{\beq}{\begin{equation}}
\newcommand{\eeq}{\end{equation}}

\newcommand{\bft}{\mathbf{t}}

\newcommand{\bfC}{\mathbf{C}}

\newcommand{\bfR}{\mathbf{R}}

\newcommand{\bfT}{\mathbf{T}}

\newcommand{\be}{\begin{equation}}
\newcommand{\ee}{\end{equation}}
\newcommand{\beqa}{\begin{eqnarray}}
\newcommand{\eeqa}{\end{eqnarray}}